\documentclass[letterpaper, 10 pt, conference]{ieeeconf}  

\IEEEoverridecommandlockouts                              

\usepackage{graphics} 
\usepackage{epsfig} 
\usepackage{times} 
\usepackage{amsmath} 
\usepackage{amssymb}  
\usepackage{mathrsfs}
\usepackage{subfigure}
\usepackage{multirow}
\usepackage{cite}
\usepackage{bm}

\usepackage{xcolor}

\title{\LARGE \bf
LOF: Structure-Aware Line Tracking based on Optical Flow
}

\author{Meixiang Quan$^{1}$, Zheng Chai$^{1}$ and Xiao Liu$^{1}$
\thanks{$^{1}$ Meixiang Quan, Zheng Chai and Xiao Liu are with Megvii (Face++) Technology Inc., Beijing, China. Email: quanmeixiang@megvii.com, icecc.sunny@gmail.com, liuxiao@foxmail.com}%
}

\begin{document}

\maketitle
\thispagestyle{empty}
\pagestyle{empty}

\begin{abstract}

Lines provide the significantly richer geometric structural information about the environment than points, so lines are widely used in recent Visual Odometry (VO) works. Since VO with lines use line tracking results to locate and map, line tracking is a crucial component in VO. Although the state-of-the-art line tracking methods have made great progress, they are still heavily dependent on line detection or the predicted line segments. In order to relieve the dependencies described above to track line segments completely, accurately, and robustly at higher computational efficiency, we propose a structure-aware Line tracking algorithm based entirely on Optical Flow (LOF). Firstly, we propose a gradient-based strategy to sample pixels on lines that are suitable for line optical flow calculation. Then, in order to align the lines by fully using the structural relationship between the sampled points on it and effectively removing the influence of sampled points on it occluded by other objects, we propose a two-step structure-aware line segment alignment method. Furthermore, we propose a line refinement method to refine the orientation, position, and endpoints of the aligned line segments. Extensive experimental results demonstrate that the proposed LOF outperforms the state-of-the-art performance in line tracking accuracy, robustness, and efficiency, which also improves the location accuracy and robustness of VO system with lines.

\end{abstract}

\section{INTRODUCTION}\label{introduction}

Visual Odometry (VO) can be used in many applications, such as augmented reality, robotics, and micro aerial vehicle. Therefore, VO has received significant interest in Robotics and Computer Vision communities. Most of the VO methods only use point features as visual information, but extracting enough reliable point features and reliably tracking them in low-texture scenes are challenging. In contrast, line features are often abundant in low-texture environments and provide the significantly richer information than point features about the environmental structure. Thus, line features have been used in recent VO works \cite{stereoSLAM, xiaojia, heyijiaplvio, structvio}. Feature tracking is a crucial component in VO process. In order to improve the performance of VO with lines, accurate and robust line tracking is required. 

Traditional line tracking algorithms can be divided into three categories: descriptor, geometry, and guided line detection-based approaches. Descriptor-based approaches \cite{MSLD}\cite{LBD} extract a sparse set of line features and compute the corresponding descriptors in each image at first, and then match the features by comparing descriptors. In general, the texture of local neighborhoods of lines are not as rich as point features, i.e., descriptors of lines are less distictive. Therefore, it is prone to give false matches in the scene with more repeated texture. In addition, it is extremely time-consuming to extract lines and calculate the corresponding descriptors for each image. Geometry-based approaches \cite{GEO1,GEO2,GEO3,KLTGEO1,KLTGEO3} extract a sparse set of line features in each image and predict each line in new image at first, and then track each line by searching for the extracted line in new image that best satisfies the corresponding geometric constraints with the predicted one. Especially, \cite{KLTGEO1}\cite{KLTGEO3} provide better predictive value by using point tracking method, such as KLT \cite{Lucas20}, to track the sampled points on line. By eliminating the need of costly descriptor computation, it can match lines faster. However, it is still time-consuming to extract lines for each image, and once the predicted line segment is not correct, it will cause false matching. Besides, image noise or occlusions often lead to the unstable line endpoints. Guided line detection-based method \cite{LineFlow} predicts each line in new image to guide the line extraction in nearby regions of the predicted one, and then select the extracted line with best collinearity as the tracked one to fuse with other extracted lines that have similar geometry property. By eliminating the need of line extraction in each image from scratch, it is more efficient in computation. In addition guided by prediction, corresponding line extraction is stable in endpoint positions even when facing temporal occlusions. However, its matching result still depends heavily on the predicted line. Although these state-of-the-art line tracking methods have made great progress, they are still heavily dependent on line detection or the predicted lines. These remaining challenges motivate us to design a line tracking algorithm that tracks lines better and faster.

\begin{figure*}[!t]
\centering
\includegraphics[width=7in]{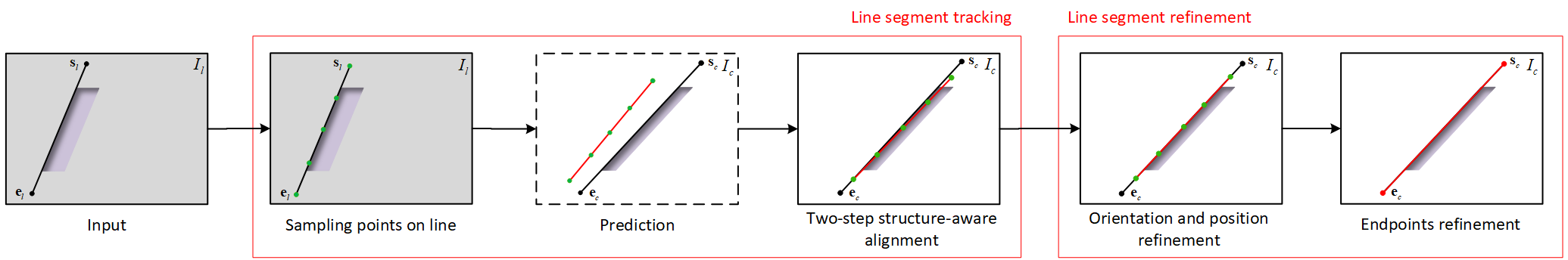}
\caption{An overview of LOF line tracking algorithm that tracks the best correspondence $\mathbf{s}$-$\mathbf{e}_c$ in image $I_c$ for line $\mathbf{s}$-$\mathbf{e}_l$ in $I_l$. Green points indicate the sampled points on line, red lines indicate the tracking results of line $\mathbf{s}$-$\mathbf{e}_l$ in corresponding steps, dotted box indicates that the step is not necessary.}
\label{fig:overview}
\end{figure*}

We propose a structure-aware Line tracking algorithm based entirely on Optical Flow (LOF), which is illustrated in Fig. \ref{fig:overview}. In order to calculate the line optical flow accurately, robustly, and efficiently, the following 5 aspects are considered in the design of the algorithm. (1) how to use the intensity values around pixels on line segment efficiently and reasonably; (2) how to calculate the line optical flow by taking each line segment as a whole, instead of independently calculating the optical flow of each sampling point on the line that ignores the structural relationship between them; (3) how to ensure the correct line optical flow calculation in the case of partial occlusion; (4) how to solve the problem of line tracking error accumulation caused by template image update; (5) how to extend the endpoints of line segment when its observation changes from short to long.

Based on the above considerations, the main contributions of this work are as follows:
\begin{itemize}
\item We propose a structure-aware line tracking algorithm based on optical flow. To the best of our knowledge, it is the first line tracking algorithm based entirely on optical flow, it also pays attention to the structural information of line segments in the process of line optical flow calculation.

\item We propose a novel line optical flow calculation method. It samples points on line by sufficiently using gradient information to ensure that the sampled points are appropriate for line optical flow calculation. Then, it performs a two-step structure-aware line alignment method to optimize the positions of sampled points on line and line parameters together, which makes full use of the structural relationship between those sampled points, and effectively removes the influence of sampled points occluded by other objects in the alignment process. (Section \ref{sec:linetracking})

\item We propose a line refinement method for the aligned lines. On the one hand, it refines the orientation and position of the aligned one to eliminate the line tracking error accumulation. On the other hand, it refines the endpoints of the aligned one to maintain complete line segments. (Section \ref{sec:linerefinement})

\item The proposed line tracking is an independent module, it can be used in any VO system. Extensive experiments were performed to demonstrate that the proposed algorithm can achieve more accurate and robust line tracking results than state-of-the-art results at lower computational cost, which also improves the localization accuracy of VO system with lines. (Section \ref{sec:experiments})
\end{itemize}

\section{Line Segment Tracking}\label{sec:linetracking}
For line segments $\mathcal{L}_l$ in last image $I_l$, LOF aims to use the intensity values around the lines to find the corresponding line segments $\mathcal{L}_c$ in current image $I_c$. We construct the image pyramid from $I_c$. Then for each line segment $\mathbf{s}$-$\mathbf{e}_l \in \mathcal{L}_l$, $\mathbf{s}$ and $\mathbf{e}$ represent the two endpoints of the line, the following steps are performed to search for its optimal correspondence $\mathbf{s}$-$\mathbf{e}_c$ in $I_c$. 

Firstly, we use strategy described in Section \ref{samplintpoint} to sample points on line segment $\mathbf{s}$-$\mathbf{e}_l$ that are appropriate for aligning the line. Then if we integrate LOF into VO system, in order to increase the possibility of line alignment converging to the global optimum, we perform prediction process described in Section \ref{prediction} to provide the better initial value for line alignment. In prediction process, the relative rotation $\mathbf{R}^c_l \in SO(3)$ from last frame to current frame can be obtained by the motion of last frame or information such as IMU and wheel encoders. Finally based on the initial value, we perform two-step structure-aware line segment alignment method described in Section \ref{stepbystepalign} to obtain the optimal correspondence $\mathbf{s}$-$\mathbf{e}_c$ in $I_c$.

\subsection{Preliminaries}
The line segment $\mathbf{s}$-$\mathbf{e}$ is represented by linear representation $\mathbf{l} = \left[ \ a \ b\ c \ \right]^\mathrm{T}$. Each pixel $\mathbf{p} = \left[ \ u \ v \ \right]^\mathrm{T}$ on the line segment $\mathbf{s}$-$\mathbf{e}$ satisfies the constraint:
\begin{equation}
\setlength{\abovedisplayskip}{2pt}
\setlength{\belowdisplayskip}{2pt}
au+bv+c=0
\label{general form}
\end{equation}

The line $\mathbf{s}$-$\mathbf{e}$ has only two degree-of-freedom, but linear representation is 3-dimensional vector, which prevents the direct application of the representation to nonlinear optimization. The equation \eqref{general form} can be expressed in normal form as:
\begin{equation}
\setlength{\abovedisplayskip}{2pt}
\setlength{\belowdisplayskip}{2pt}
\mathrm{cos}\beta u + \mathrm{sin}\beta v - d = 0
\label{normal form}
\end{equation}
where $\beta \in [ 0, \pi )$ is the angle from the x-axis of image coordinates to the normal line segment that is drawn from the origin of image coordinates perpendicular to the line, and $d \geq 0$ is the length of the normal line segment.

Therefore, we can use normal representation $\mathbf{U} = [ \ \beta \ d \ ]^\mathrm{T}$ to represent the line segment during optimization. From \eqref{general form} and \eqref{normal form}, we can know that linear representation and normal representation can be converted to each other:
\begin{equation}
\setlength{\abovedisplayskip}{2pt}
\setlength{\belowdisplayskip}{2pt}
\beta = \mathrm{arctan} \left( \frac{b}{a} \right), \ d = - \frac{c}{\sqrt{a^2 + b^2}}
\label{linear2normal}
\end{equation}
\begin{equation}
\setlength{\abovedisplayskip}{2pt}
\setlength{\belowdisplayskip}{2pt}
a = \mathrm{cos} \beta, \ b = \mathrm{sin} \beta, \ c = -d
\label{normal2linear}
\end{equation}

The angle from the x-axis of image coordinates to line segment $\mathbf{s}$-$\mathbf{e}$ is computed as:
\begin{equation}
\setlength{\abovedisplayskip}{2pt}
\setlength{\belowdisplayskip}{2pt}
\alpha = \mathrm{arctan}\left( -\frac{a}{b} \right)
\end{equation}

For pixel $\mathbf{p}$, its gradient magnitude $G$ and the angle $\theta$ from the x-axis of image coordinates to the line perpendicular to its gradient direction are computed as:
\begin{equation}
\setlength{\abovedisplayskip}{2pt}
\setlength{\belowdisplayskip}{2pt}
G = \sqrt{g_u^2 + g_v^2} , \ \theta = \mathrm{arctan}(- \frac{g_u}{g_v})
\end{equation}
where $\nabla I = \left[ \ g_u \ g_v \ \right]$ is the gradient of image $I$ at pixel $\mathbf{p}$. In addition, the normal vector $\mathbf{n}$ of line segment $\mathbf{s}$-$\mathbf{e}$ is computed as:
\begin{equation}
\setlength{\abovedisplayskip}{2pt}
\setlength{\belowdisplayskip}{2pt}
\mathbf{n} = \left[\begin{array}{cc} \frac{a}{\sqrt{a^2+b^2}} & \frac{b}{\sqrt{a^2+b^2}} \end{array} \right]^\mathrm{T}
\end{equation}

\subsection{Sampling Points on Line}\label{samplintpoint}
For each line segment $\mathbf{s}$-$\mathbf{e}_l \in \mathcal{L}_l$, we evenly sample $n$ points $\mathcal{P}_l = \{\mathbf{p}_{l_1}, \mathbf{p}_{l_2}, ..., \mathbf{p}_{l_n} \}$. Then, we judge whether the sampled points $\mathcal{P}_l$ are appropriate for aligning the line segment $\mathbf{s}$-$\mathbf{e}_l$. The sampled point $\mathbf{p}_l \in \mathcal{P}_l$ is considered appropriate when the following condition is met:
\begin{equation}
\setlength{\abovedisplayskip}{2pt}
\setlength{\belowdisplayskip}{2pt}
|\alpha_l - \theta_l| < T_\theta, \ G_l > T_G
\label{criteria}
\end{equation}
where $\alpha_l$ denotes the orientation of line $\mathbf{s}$-$\mathbf{e}_l$, $G_l$ and $\theta_l$ denote the gradient magnitude and orientation of sampled point $\mathbf{p}_l$, $T_\theta$ and $T_G$ denote the corresponding threshold.

That is, we only consider those sampled points with obvious gradient change along the direction of normal vector of line segment $\mathbf{s}$-$\mathbf{e}_l$ to be appropriate for aligning the line. Let $\mathcal{P}_l^{'}$ represent the sampled points in $\mathcal{P}_l$ that do not satisfy condition \eqref{criteria}. To remedy some of the sampled points in $\mathcal{P}_l^{'}$, we move each sampled point $ \mathbf{p}_l^{'} \in\mathcal{P}_l^{'}$ by $k$ pixels along the line direction to obtain a new pixel, where we determine whether condition \eqref{criteria} is satisfied. If it is met, we use the new pixel to replace the original sampled point $\mathbf{p}_l^{'}$, else we remove the sampled point $\mathbf{p}_l^{'}$ from $\mathcal{P}_l$. 

In this way, we can exclude the sampled points on poor texture areas or occluding objects that have the negative impact on the alignment of line segment $\mathbf{s}$-$\mathbf{e}_l$, so that only the sampled points strictly belonging to the line segment are used for its alignment.

\subsection{Prediction}\label{prediction}
If two images $I_l$ and $I_c$ are assumed to be taken by a pure camera rotation $\mathbf{R}^c_l$, the motion of sampled points $\mathcal{P}$ and line segment $\mathbf{s}$-$\mathbf{e}$ between the two images can be described by 2D homography $\mathbf{H}^c_l$:
\begin{equation}
\setlength{\abovedisplayskip}{2pt}
\setlength{\belowdisplayskip}{2pt}
\mathbf{H}^c_l = \mathbf{K} \mathbf{R}^c_l \mathbf{K}^{-1}
\end{equation}
where $\mathbf{K}$ is the camera intrinsic matrix. Hence, we can use the predicted relative rotation $\mathbf{R}^c_l$ to predict the initial value of each sampled point $\mathbf{p}_l \in \mathcal{P}_l$ and linear representation $\mathbf{l}_l$ in $I_c$ as:
\begin{equation}
\setlength{\abovedisplayskip}{2pt}
\setlength{\belowdisplayskip}{2pt}
\hat{\mathbf{p}}_c = \mathbf{H}^c_l \mathbf{p}_l, \ \hat{\mathbf{l}}_c = {\mathbf{H}^c_l}^{-\mathrm{T}} \mathbf{l}_l\\
\end{equation}

\subsection{Two-Step Structure-Aware Alignment}\label{stepbystepalign}

\subsubsection{Distinction Between Corner and Edge-like Points} \label{distinction}
For each corner-like point $\mathbf{p}_l$ in $I_l$, 2D feature alignment optimizes for the correction $\bm{\delta} \mathbf{p}_c \in \mathbb{R}^2$ to obtain the corresponding optimal position $\mathbf{p}_c$ in $I_c$, i.e. $\mathbf{p}_c \! \leftarrow \! \mathbf{p}_c + \bm{\delta} \mathbf{p}_c$. However, if the 2D feature alignment is performed for edge-like points, the points may drift along the edge. Therefore, in order to align sampled points $\mathcal{P}_l$ on line $\mathbf{s}$-$\mathbf{e}_l$ along the optimal direction, we classify the sampled points according to the eigenvalue of their gradient matrix. If the minimum eigenvalue is large enough, we think it's a corner-like point, and if one eigenvalue is large enough and the other is small, we think it's a edge-like point. Then for each edge-like point $\mathbf{p}_l$ in $I_l$, we optimize for correction $\delta p_c \in \mathbb{R}^1$ in the direction of the edge normal $\mathbf{n}_c$ to obtain the corresponding optimal position $\mathbf{p}_c$ in $I_c$, i.e. $\mathbf{p}_c \! \leftarrow \! \mathbf{p}_c + \delta p_c \mathbf{n}_c$. 

\subsubsection{Structure-Aware Alignment}\label{structureawarealingment}
Let $\mathcal{P}_l^\mathcal{C} \subseteq \mathcal{P}_l$ and $\mathcal{P}_l^\mathcal{E} \subseteq \mathcal{P}_l$ represent the corner-like and edge-like sampled points in $\mathcal{P}_l$ respectively. In the line alignment process, we want to make full use of the structural relationship between the sampled points on line, and optimize the line parameters to provide the better optimization direction for edge-like points. Therefore, we align the line segment $\mathbf{s}$-$\mathbf{e}_l$ by jointly optimizing the positions of the sampled points $\mathcal{P}_l$ in $I_c$ and normal representation $\mathbf{U}_c$ as follows:
\begin{equation}
\setlength{\abovedisplayskip}{2pt}
\setlength{\belowdisplayskip}{2pt}
\bm{\mathcal{X}} = \{ \mathcal{P}_c, \mathbf{U}_c \}
\end{equation}
\begin{equation}
\setlength{\abovedisplayskip}{2pt}
\setlength{\belowdisplayskip}{2pt}
\begin{split}
\bm{\mathcal{X}} &= \mathop{\mathrm{argmin}}_{\bm{\mathcal{X}}} \sum_{ \mathbf{p}_c \in \mathcal{P}_c} \sum_{\mathbf{h} \in \mathcal{W}} {\left\| T_l(\mathbf{h}) - I_c \left( \mathbf{p}_c + \mathbf{h} \right) \right\|}^2 \\
&+ \sum_{\mathbf{p}_c \in \mathcal{P}_c} {\left\| [ \mathrm{cos}\beta_c \ \mathrm{sin}\beta_c] \cdot \mathbf{p}_c - d_c \right\|}^2
\end{split}
\label{linealignment}
\end{equation}
where $\mathcal{P}_c$ is the correspondence of $\mathcal{P}_l$ in $I_c$, the first error term is the photometric error of the corner-like points $\mathcal{P}_l^\mathcal{C}$ and edge-like points $\mathcal{P}_l^\mathcal{E}$ between two images, and the second error term is the structural error of the sampled points deviating from the line. Besides, $\mathcal{W} = \{\mathbf{h} | -10 \leq h_u \leq 10, \ -10 \leq h_v \leq 10 \}$ and
\begin{equation*}
\setlength{\abovedisplayskip}{2pt}
\setlength{\belowdisplayskip}{2pt}
T_l(\mathbf{h}) = I_l \left( \mathbf{p}_l + {\mathbf{A}^c_l}^{-1} \mathbf{h} \right), \mathbf{A}^c_l = \frac{\mathbf{H}^c_l}{h_{33}}|_{2\times 2}
\end{equation*}
where $2\times 2$ denotes upper $2\times 2$ matrix. 

Based on the initial value $\hat{\bm{\mathcal{X}}}$, the optimization problem in \eqref{linealignment} is solved by the inverse compositional algorithm \cite{inversecompositional}. The inverse compositional algorithm iteratively solves for the increments $\delta \bm{\mathcal{X}}$, and updates the parameters as:
\begin{equation}
\setlength{\abovedisplayskip}{2pt}
\setlength{\belowdisplayskip}{2pt}
\begin{split}
\mathbf{p}_c &\leftarrow \left\{\begin{array}{cc} \hat{\mathbf{p}}_c + \bm{\delta} \mathbf{p}_c, & \mathbf{p}_l \in \mathcal{P}_l^\mathcal{C} \\ \hat{\mathbf{p}}_c + \left[\begin{array}{c} \mathrm{cos}\hat{\beta}_c \\ \mathrm{sin}\hat{\beta}_c \end{array} \right] \delta p_c, & \mathbf{p}_l \in \mathcal{P}_l^\mathcal{E} \end{array}\right.\\
\beta_c &\leftarrow \hat{\beta}_c + \delta \beta_c\\
d_c &\leftarrow \hat{d}_c + \delta d_c
\end{split}
\end{equation}

\subsubsection{Two-Step Alignment}\label{twostepalignment}
The first step of alignment is performed as described in Section \ref{structureawarealingment} by using all the sampled points $\mathcal{P}_l$. The first step alignment is terminated when the following convergence condition is satisfied by more than 40\% of sampled points in one iteration. We think the alignment of sampled point $\mathbf{p}_l \in \mathcal{P}_l$ in $I_c$ is converged when its update is lower than a given threshold, and the aligned point $\mathbf{p}_c$ and line $\mathbf{l}_c$ satisfy the condition \eqref{criteria}. After the first step alignment, we use the converged sampling points $\mathcal{P}^{(c)}_{l}$ to perform the second step of line segment alignment. The sampled points on line with slower convergence are likely to be those on the occluding objects, which are inconsistent with the converged points on line and line parameters. Therefore, the second step alignment can effectively avoid the negative impact of those points with slower convergence on the line segment alignment. When the positions of the sampled points $\mathcal{P}^{(c)}_{l}$ in $I_c$ and the line parameters $\mathbf{U}_c$ are all converged, we think the second step alignment is successed. Then for the sampled points in $\mathcal{P}_l$ that fails to converge, we project its initial value in $I_c$ onto the aligned line $\mathbf{l}_c$.

\subsubsection{Pyramidal Implementation}

In this work, we perform the pyramidal line segment alignment. At each pyramid level $L$, we firstly remove those points located at the boundary of $I_l^L$ or $I_c^L$ from the sampled points $\mathcal{P}_l^L$ to construct the available sampled points $\mathcal{P}_l^{L''}$. At the original image level, we further remove those sampled points whose minimum eigenvalue of gradient matrix \cite{goodfeaturestotrack} is too high, which avoids the negative influence of sampled points near occluding objects on line alignment. Next, we classify the sampled points $\mathcal{P}_l^{L''}$ into corner-like points and edge-like points according to the method described in Section \ref{distinction}. Then by using the available sampled points $\mathcal{P}_l^{L''}$, we perform the two-step line segment alignment method described in Section \ref{twostepalignment} to align the line segment $\mathbf{s}$-$\mathbf{e}_l^L$ in image $I_c^L$.

The overall pyramidal line segment alignment algorithm proceeds as follows. First, the line optical flow is computed at the deepest pyramid level $L$. Then, the result of that computation is propagated to the upper level $L-1$ as initial guess. Let $s$ be the pyramid scale, the line parameters are then propagated as $\beta_c^{L-1} = \beta_c^L$ and $d_c^{L-1} = s \cdot d_c^L$. Given that initial guess, the refined line optical flow is comuputed at uptter level, and the result is propagated to level $L-2$ and so on up to the original image level. At the original image level, we take all the sampled points in $\mathcal{P}_c^{''}$ that satisfy the condition \eqref{criteria} with tracked line $\mathbf{l}_c$ as the successfully traked points on line, denoted by $\mathcal{P}_{c (c)}^{''}$.

\section{Line Segment Refinement}\label{sec:linerefinement}
The tracked line segment may be not correctly the real one due to noise or occluding object. So after line segment alingment, we refine the tracked line segment $\mathbf{s}$-$\mathbf{e}_c \in \mathcal{L}_c$ to better fit the real one by using intensity information. Firstly, we refine the orientation and position of the tracked line segment as described in Section \ref{linearrefine}. Then, we refine the endpoints of the tracked line segment as described in Section \ref{endpointsrefine} to maintain the complete line segment in each image, not just a part of the line segment.

\subsection{Orientation and Position Refinement}\label{linearrefine}
For all the tracked sampling points $\mathcal{P}_{c (c)}^{''}$, we compute their photometric error between $I_l$ and $I_c$, in which we select the point $\mathbf{p}_{c}^{\ast}$ with the smallest photometric error. Next, we adjust the position of tracked line $\mathbf{l}_c$, i.e., $d_c$ in $\mathbf{U}_c$, so that it passes through point $\mathbf{p}_{c}^{\ast}$, and then project all the points in $\mathcal{P}_{c (c)}^{''}$ to the line $\mathbf{l}_c$. Let $g = |\alpha_c - \hat{\alpha}_c|$ represents the angle difference between the tracked and predicted line segment. We then take the point $\mathbf{p}_{c}^{\ast}$ as the center to rotate the tracked line segment by $[-g, g]$ degrees in $N=\mathrm{min}(20, 20 * g)$ steps. For each rotated line segment $\mathbf{s}$-$\mathbf{e}_c^{r}$ with parameter $\mathbf{l}_c^{r}$, we compute the sum $\Sigma$ of the gradients of the rotated sampling points $\mathcal{P}_{c (c)}^{r''}$  along the normal vector $\mathbf{n}_c^{r}$ of the line as:
\begin{equation}
\setlength{\abovedisplayskip}{2pt}
\setlength{\belowdisplayskip}{2pt}
\Sigma = \sum_{\mathbf{p}_{c (c)}^{r''} \in \mathcal{P}_{c (c)}^{r''}} \left| \frac{I_c(\mathbf{p}_{c (c)}^{r''} + \mathbf{n}_c^{r})- I_c(\mathbf{p}_{c (c)}^{r''} - \mathbf{n}_c^{r})}{2} \right|
\label{gradientsum}
\end{equation}

Among all the rotated line segments, we think the line segment with the most obvious gradient change along the normal vector of the line is the best line segment to fit the real one. Therefore, we select the rotated line segment $\mathbf{s}$-$\mathbf{e}_c^r$ with highest $\Sigma$ as the refined one $\mathbf{s}$-$\mathbf{e}_c$.

\subsection{Endpoints Refinement}\label{endpointsrefine}
Among the refined sampling points $\mathcal{P}_{c (c)}^{''}$, the two most distant points $\mathbf{p}_c^{s}$ and $\mathbf{p}_c^{e}$ form the endpoints of the refined line segment $\mathbf{s}$-$\mathbf{e}_c$ in $I_c$. Then, we extend the two line endpoints recursively to maintain the complete line segment. For endpoint $\mathbf{p}_c^{s}$, the pixel $\mathbf{p}_c^{s'}$ near to it in the opposite direction of the line is tested. If the gradient of pixel $\mathbf{p}_c^{s'}$ satisfies the condition \eqref{criteria} with the refined line $\mathbf{l}_c$, we set the pixel $\mathbf{p}_c^{s'}$ as new endpoint, and continue to extend it until the condition \eqref{criteria} is no longer satisfied. The endpoint $\mathbf{p}_c^{e}$ is extended in same way as endpoint $\mathbf{p}_c^{s}$.

\begin{table*}[t]
	\caption{Line tracking performance on TUM RGBD dataset. The best results are given in \textbf{bold}.}
	\label{tab:tum}
	\begin{center}
		\begin{tabular}{c| c c c| c c c | c c c}
			\hline
			\hline
			\multirow{3}{*}{Sequence}
			&\multicolumn{6}{c|}{Track Number \& Accuracy Test} & \multicolumn{3}{c}{Track Length Test} \\ \cline{2-10}
			&\multicolumn{3}{c|}{Number of Matches} & \multicolumn{3}{c|}{Matching Accuracy} & \multicolumn{3}{c}{Tracking Length} \\ 
			& LBD\cite{LBD} & LGK\cite{KLTGEO3} & LOF & LBD\cite{LBD} & LGK\cite{KLTGEO3} & LOF & LBD\cite{LBD} & LOF w/o ori. pos. refine & LOF \\ \cline{1-10}
			\texttt{fr1\_360} & 53 & 48 & \textbf{62} & 80\% & \textbf{94}\% & 93\%  & 5.3 & 8.2 & \textbf{8.5}  \\
			\texttt{fr1\_desk} & 68 & 56 & \textbf{77} & 80\% & 94\% & \textbf{95}\%  & 6.5 & 16.4 & \textbf{17.5} \\
			\texttt{fr1\_floor} & 57 & 57 & \textbf{75} & 83\% & \textbf{99}\% & 97\%  & 6.8 & 22.0 & \textbf{24.6} \\
			\texttt{fr1\_room} & 69 & 60 & \textbf{80} & 81\% & \textbf{95}\% & 94\%  & 6.7 & 16.1 & \textbf{17.4} \\
			\texttt{fr2\_360\_hemisphere} & 78 & 70 & \textbf{79} & 87\% & 96\% & \textbf{98}\%  & 9.0 & 25.5 & \textbf{27.7} \\
			\texttt{fr2\_360\_kidnap} & 89 & 66 & \textbf{90} & 87\% & 95\% & \textbf{99}\%  & 10.8 & 44.0 & \textbf{51.6} \\
			\texttt{fr2\_large\_with\_loop} & \textbf{86} & 75 & \textbf{86} & 87\% & 96\% & \textbf{98}\%  & 8.9 & 25.6 & \textbf{28.1} \\
			\texttt{fr3\_long\_office} & 74 & - & \textbf{84} & 91\% & - & \textbf{99}\%  & 12.5 & 79.7 & \textbf{94.0} \\
			\texttt{fr2\_pioneer\_slam3} & 45 & - & \textbf{52} & 78\% & - & \textbf{93}\%  & 7.0 & 15.1 & \textbf{16.2} \\
			\texttt{fr3\_str\_notex\_far} & 25 & - & \textbf{33} & 92\% & - & \textbf{99}\%  & 5.7 & 124.7 & \textbf{152.4} \\
			\texttt{fr2\_desk\_with\_person} & 49 & - & \textbf{55} & 85\% & - & \textbf{98}\%  & 11.1 & 68.5 & \textbf{70.3} \\
			\texttt{fr3\_sitting\_static} & 84 & - & \textbf{89} & 90\% & - & \textbf{100}\%  & 14.9  & 65.3 & \textbf{67.5}\\
			\texttt{fr3\_sitting\_xyz} & 76 & - & \textbf{84} & 85\% & - & \textbf{97}\%  & 10.5 & 126.3 & \textbf{150.6} \\			
			\texttt{fr3\_walking\_half} & 71 & - & \textbf{79} & 80\% & - & \textbf{93}\%  & 6.9 & 25.0 & \textbf{26.8} \\
			\texttt{fr3\_walking\_rpy} & 67 & - & \textbf{75} & 76\% & - & \textbf{93}\%  & 5.2 & 15.5 & \textbf{16.3} \\
			\hline
			\texttt{average} & 66 & 61 & \textbf{73} & 84\% & 95\% & \textbf{96}\% & 8.5 & 45.2 & \textbf{51.3} \\
			\hline
			\hline
		\end{tabular}
	\end{center}
\end{table*}

\section{Experiments}\label{sec:experiments}
We evaluate our line tracking algorithm LOF, and compare it with the descriptor-based algorithm LBD \cite{LBD} and geometry-based algorithm LGK \cite{KLTGEO3} on two public datasets: TUM RGBD \cite{tum} and EuRoC\cite{EUROC}. LBD is open-source, so we can make a fair comparison in all experiments. LGK \cite{KLTGEO3} shows its line tracking results on TUM RGBD dataset, so allowing for a direct comparison in some experiments. Firstly, we evaluate the line tracking performance in Section \ref{linetrackingperformance}. TUM RGBD dataset provides the depth information and groundtruth transformation, which gives the correct line matching relationship between two images. Therefore, the quantitative analysis for line tracking performance is carried out on TUM RGBD dataset. We think each match to be correct if the correspondent line projection error is less than 5 pixels \cite{GEO3}\cite{KLTGEO3}. Then in Section \ref{VOevaluation}, we integrate LOF into PL-VIO system \cite{heyijiaplvio} to further evaluate the line tracking performance on EuRoC. In PL-VIO, we maintain 50 lines for each frame. We use open-source tool evo \cite{evo} to evaluate the absolute pose error (APE) between the estimated trajectory and the groundtruth. All the experiments were performed on a laptop with an Intel Core i7-10510U CPU with 1.80GHz and 16GB RAM.

\subsection{Implementation Details}
The image pyramid is constructed with down-sampling scale $s=1.5$, and then pyramid height is set to 4. The thresholds in LOF are set as  $T_G = 5$, $T_\theta = 22.5^{\circ}$. All the line tracking algorithms are used with a combination to LSD line detector \cite{LSD}. LSD line extraction and LBD line matching are implemented with OpenCV 3 library\cite{opencv}. 

\subsection{Line Tracking Performance}\label{linetrackingperformance}

We perform two experiments to evaluate the line tracking performance. In the first experiment called Track Number \& Accuracy Test, we extract 100 lines in each image of sequence, and then track them in the next consecutive image, which evaluates the capability for accurate tracking between two images. In the second experiment called Track Length Test, we track the existing lines in last frame at first, and then if there are lines not successfully tracked, we extract new lines to mantain $N_l=50$ lines for each image, which evaluates the capability for a long-frame tracking.

In Table \ref{tab:tum}, we show the line tracking performance of LBD, LGK, and LOF on TUM RGBD dataset, where "$-$" indicates that the method does not give the result on the sequence. The number of matches refers to the average number of lines that are successfully matched between two images, matching accuracy refers to the ratio of the number of correct matches to the number of matches, and tracking length refers to the average length of correct tracking for each line. From the result, we can draw the following conclusions. Compared with LBD, LOF achieves better line tracking performance in the number of matches, matching accuracy, and tracking length. The improvements are because (1) LOF is a problem of solving the optimal solution, which increases the possibility of successful line tracking; (2) LOF samples points on line that are appropriate for line optical flow calculation to perform two-step structure-aware line alignment process, which can increase the possibility of correct line tracking; (3) LOF performs the pyramidal line alignment, which increases the possibility that LOF converges to the global optimum, that is, the possibility of correct line tracking; (4) LOF can turn some wrongly tracked line segments into correct one by refining the orientation and position of the tracked line segments, which can eliminate the line tracking error accumulation caused by template patch update to enable the correct line tracking for a long-frame. Then compared with LGK, although the matching accuracy of LOF is sometimes slightly lower, the number of correct matches of LOF obtained after the number of matches multiplied by matching accuracy is still higher. LGK tracks lines in two steps: line prediction and refinement. The line prediction method of LGK is based on optical flow, which corresponds to the line tracking process of LOF. It samples points for each extracted line, tracks them with IMU-aided KLT, and then use the tracked sampling points to fit a line. Compared with LOF that makes full use of the structural relationship between the sampled points on line, LGK ignores the relationship. Thus, although the line refinement method of LGK that uses the extracted lines is better than the line refinement method of LOF that only uses intensity information, the line tracking performance of LOF is still better than LGK. It can demonstrate the superiority of line alignment method of LOF shown in \eqref{linealignment}. In addition, the comparison between LOF w/o ori. pos. refine and LOF can prove the effectiveness of orientation and position refinement method of LOF. When the refinement process is removed from LOF, the line tracking error is accumulated, so the tracked line will drift over time, which shortens the length of correct line tracking. 


\begin{figure}[t]
	\centering
	\subfigure[\texttt{LOF}]{		
	\includegraphics[width=\columnwidth]{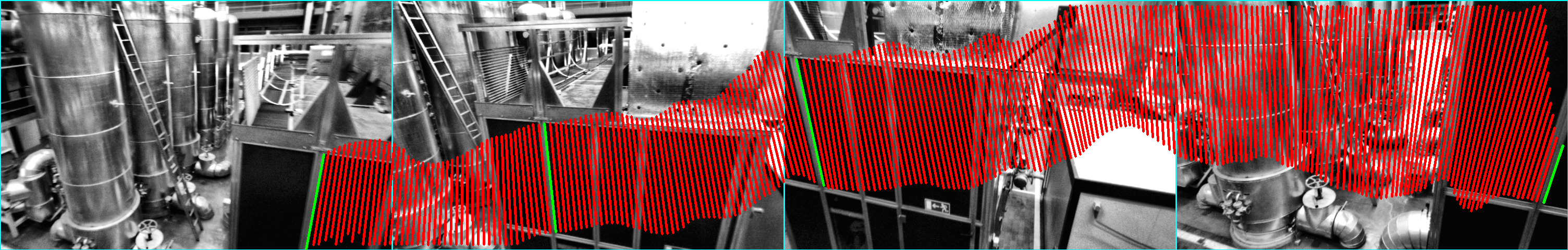}
	\label{fig:lineopticalflowa}
	}
	\subfigure[\texttt{LOF w/o endpoints refine}]{		
	\includegraphics[width=\columnwidth]{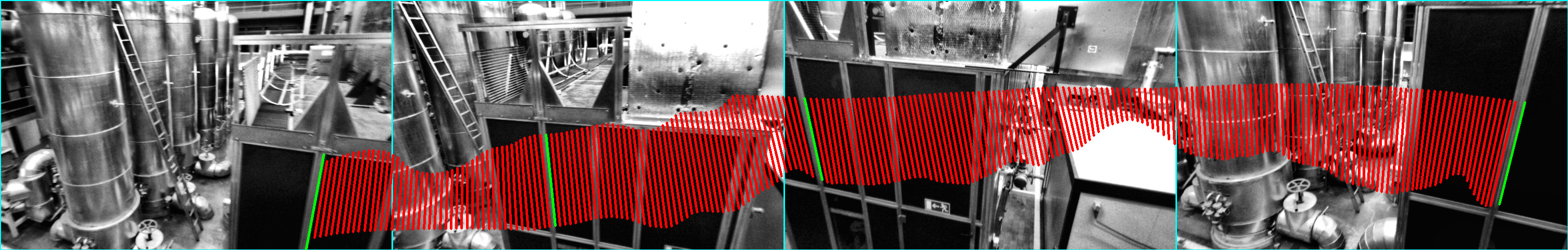}
	\label{fig:lineopticalflowb}
	}
	\subfigure[\texttt{LOF}]
	{	
		\includegraphics[width=0.7\columnwidth]{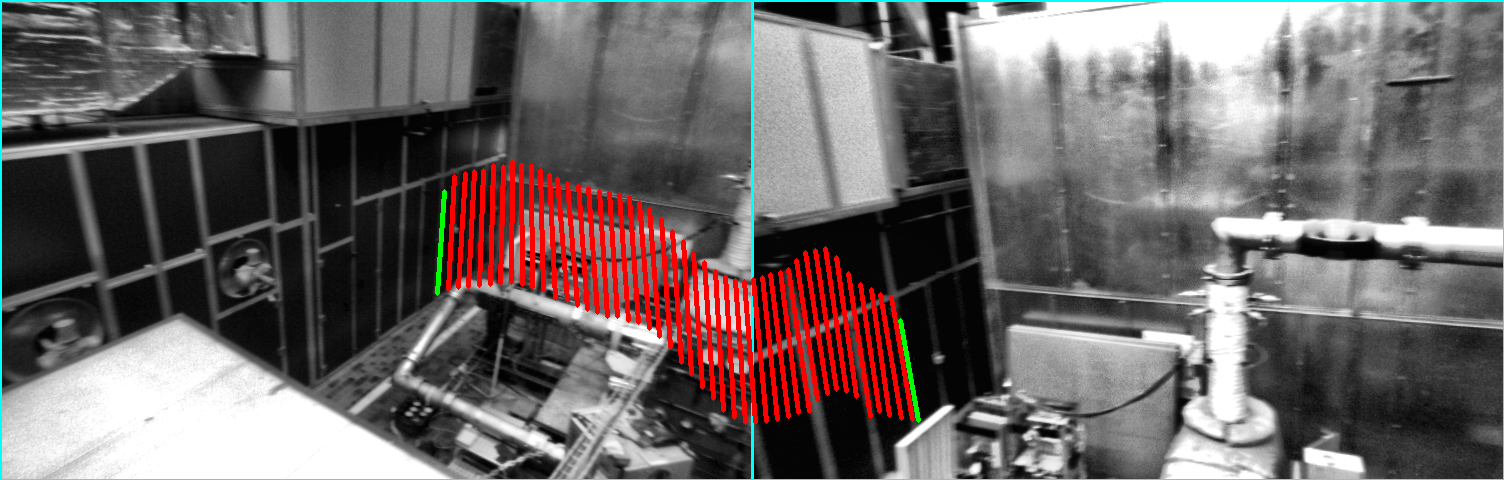}
		\label{fig:lineopticalflowc}  
	}
	\subfigure[\texttt{LOF w/o two-step align}]
	{		\includegraphics[width=0.7\columnwidth]{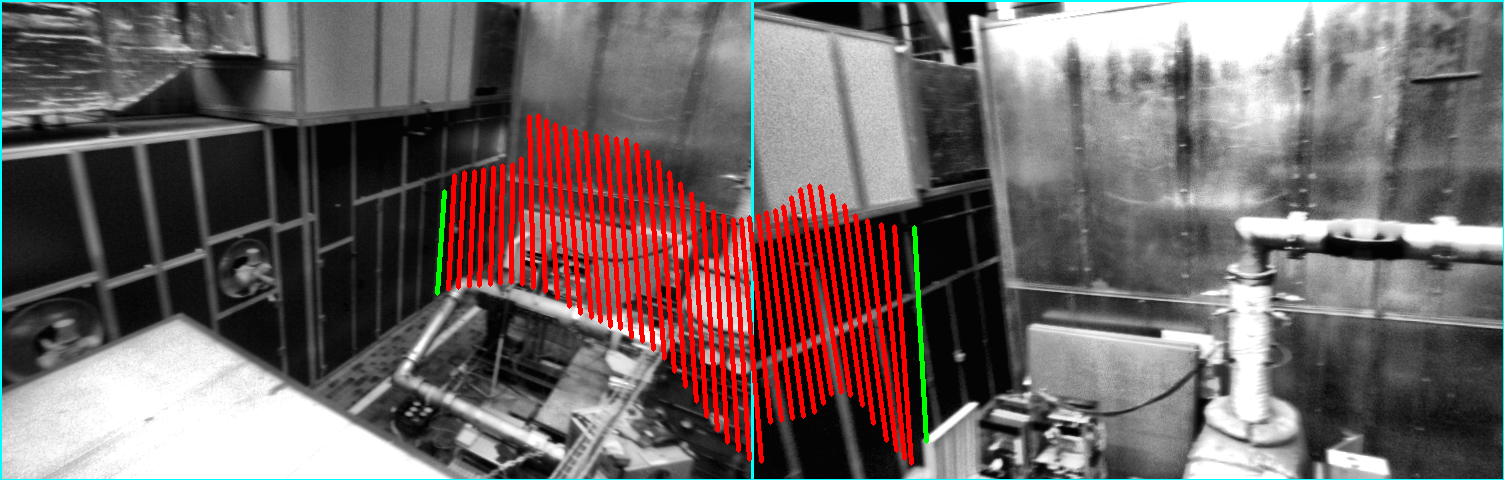}
	\label{fig:lineopticalflowd}  
	}
	\caption{Tracking results of a line from its extraction to final tracking on MH\_01\_easy (a)(b), MH\_03\_medium (c)(d) sequences. Green lines are in the corresponding images, and red lines are the tracking results in all undisplayed images.}
	\label{fig:lineopticalflow}
\end{figure}

In Fig. \ref{fig:lineopticalflow}, we show the tracking results of lines from its extraction to final tracking. Fig. \ref{fig:lineopticalflowa} demonstrates that under different viewpoints caused by camera movement, LOF can track line segments completely without drift for a long frame. By comparing Fig. \ref{fig:lineopticalflowa} and Fig. \ref{fig:lineopticalflowb}, we can prove the effectiveness of the proposed line endpoints extension method in maintenance of the complete line segments. Then, Fig. \ref{fig:lineopticalflowc} demonstrates that LOF can effectively deal with partial occlusion. By comparing Fig. \ref{fig:lineopticalflowc} and Fig. \ref{fig:lineopticalflowd}, we can verify the effectiveness of the proposed two step line alignment method in dealing with partial occlusion, which can effectively remove the influence of sampled points occluded by other objects.


\begin{figure}[!t]
\centering
\includegraphics[width=0.6\columnwidth]{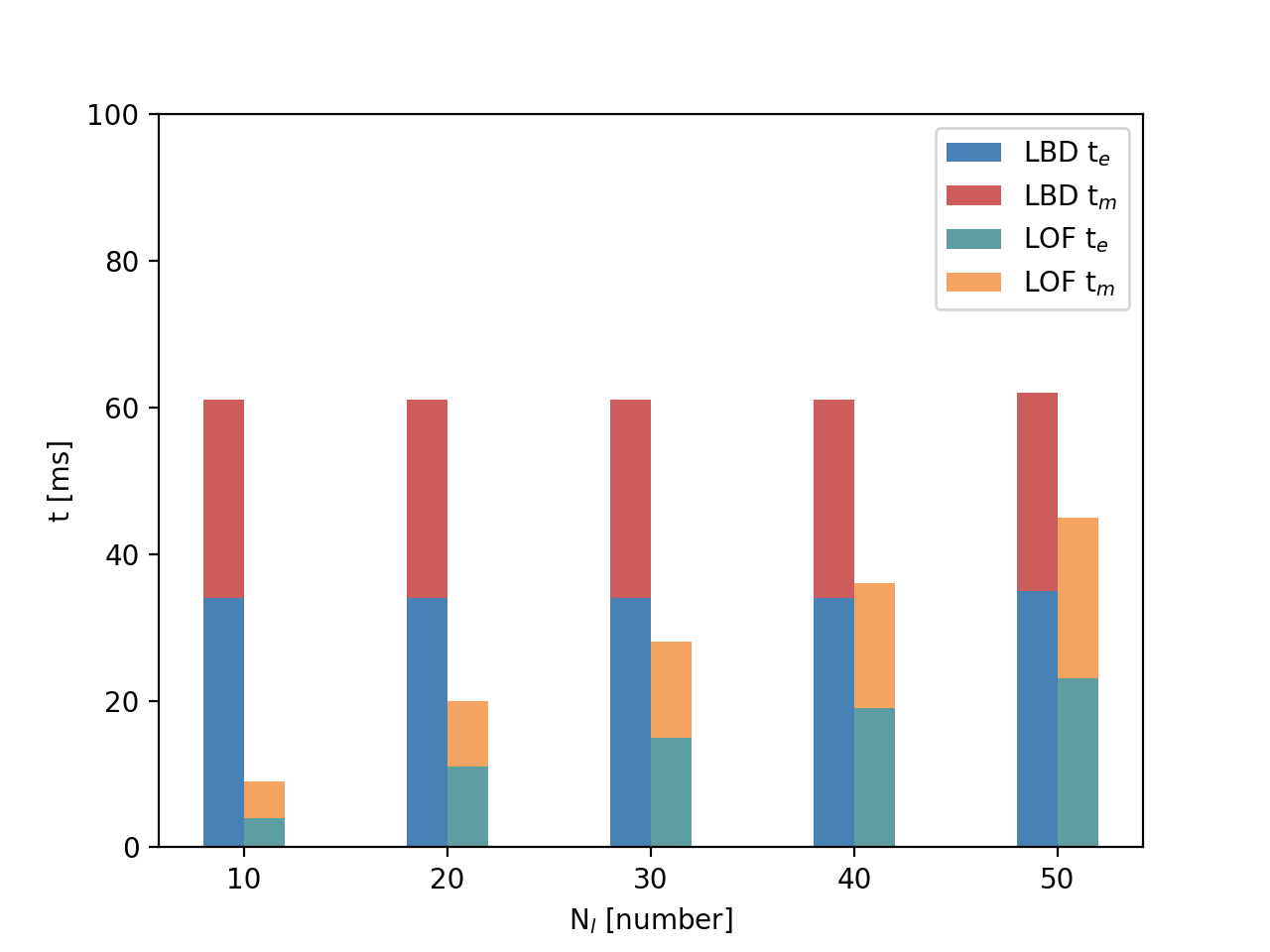}
\caption{Comparison of computational cost when performing Track Length Test experiment on fr3\_walking\_xyz sequence. The average runtime of line extraction $t_e$ and line tracking $t_m$ under different number $N_l$ of lines maintained in each frame.}
\label{fig:computationalcost}
\end{figure}

Fig. \ref{fig:computationalcost} compares the computational cost of LBD and LOF in Track Length Test experiment. Obviously, LOF tracks lines faster than LBD, especially compared with LBD that takes 27ms in average, LOF only takes 5ms when $N_l = 10$. It is because LOF directly operates on the raw pixel intensities of image to eliminate the need of costly descriptor computation. Then, the result also shows that as $N_l$ decreases from 50 to 10, the average line extraction time $t_e$ remains unchanged at 34 ms when LBD is used for line tracking, but $t_e$ decreases from 23 ms to 4 ms when LOF is used. This is because compared with LBD line matching method that needs to extract line features for each frame no matter what $N_l$ is, LOF line tracking method extracts line features only if there are lines not successfully tracked. 

\subsection{Evaluation on Visual-Inertial Odometry}\label{VOevaluation}

\begin{table}[t]
	\caption{RMSE of PL-VIO system with different line tracking methods on EuRoC dataset. The best results are given in \textbf{bold}. (Trans.: translation error (cm), Rot.: rotation error (deg))}
	\label{tab:PLVIORMSE}
	\begin{center}
		\begin{tabular}{c| c c | cc | c c}
			\hline
			\hline
			\multirow{2}{*}{Sequence}
			&\multicolumn{2}{c|}{LBD\cite{LBD}} & \multicolumn{2}{c|}{LOF} & \multicolumn{2}{c}{G-LOF} \\
			& Trans. & Rot. & Trans. & Rot. & Trans. & Rot.\\
			\hline
			\texttt{MH\_01} & 16.7 & 1.4 & 16.3 & \textbf{1.3}  & \textbf{15.1} & \textbf{1.3}  \\
			\texttt{MH\_02} & \textbf{14.2} & \textbf{1.8} & 15.8 & 2.0  & 15.7 & 2.0  \\
			\texttt{MH\_03} & 26.2 & 1.6 & 24.1 & 1.6 & \textbf{23.9} & \textbf{1.5}  \\
			\texttt{MH\_04} & 35.9 & \textbf{1.5} & \textbf{30.2} & \textbf{1.5}  & 31.1 & 1.6  \\
			\texttt{MH\_05} & 29.2 & 1.1 & \textbf{26.6} & \textbf{0.9}  & 26.7 & \textbf{0.9}  \\
			\texttt{V1\_01} & 7.4 & 1.0 & \textbf{6.1} & \textbf{0.8}  & 6.4 & 0.9  \\
			\texttt{V1\_02} & \textbf{8.7} & \textbf{2.1} & 9.2 & 2.2  & 9.6 & 2.2  \\
			\texttt{V1\_03} & 27.7 & 4.6 & \textbf{21.9} & 3.1 & 22.0 & \textbf{3.0}  \\
			\texttt{V2\_01} & 8.7 & 2.0 & \textbf{8.3} & \textbf{1.9}  & 9.5 & 2.0  \\
			\texttt{V2\_02} & 14.8 & \textbf{2.7} & 14.2 & 2.9  & \textbf{13.1} & 3.0  \\
			\texttt{V2\_03} & \textbf{32.4} & \textbf{4.1} & 36.9 & 4.2  & 34.0 & 4.4  \\
			\hline
			\texttt{average} & 20.2 & 2.2 & 19.0 & \textbf{2.0} & \textbf{18.8} & 2.1  \\
			\hline
			\hline
		\end{tabular}
	\end{center}
\end{table}

\begin{figure}[t]
	\centering
	\subfigure[\texttt{LOF}]{		
	\includegraphics[width=0.9\columnwidth]{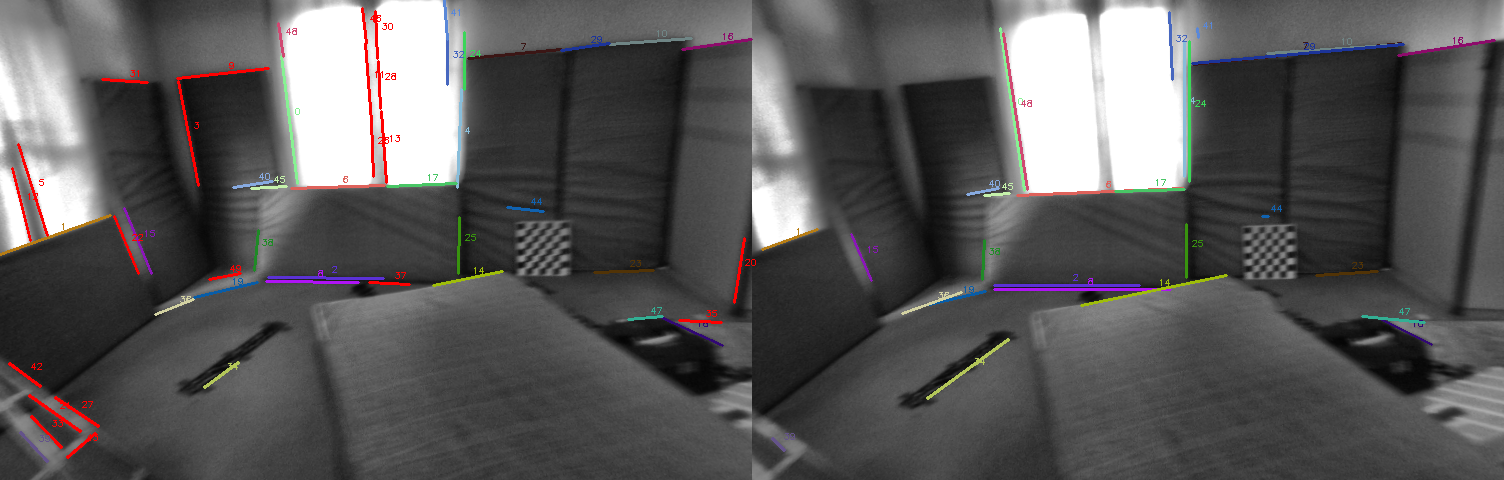}
	\label{fig:linematchinga}
	}
	\subfigure[\texttt{G-LOF}]
	{	
		\includegraphics[width=0.9\columnwidth]{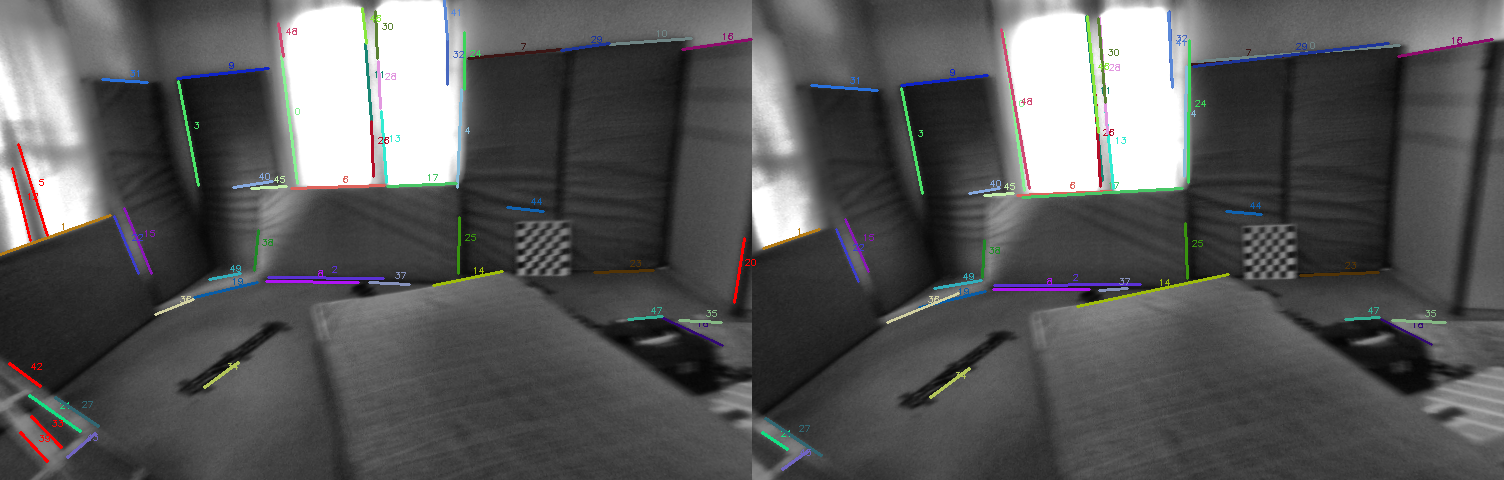}
		\label{fig:linematchingb}  
	}
	\caption{Line tracking results between two consecutive images from V1\_03\_difficult sequence. Red lines indicate that the correspondences are not found. Obviously, G-LOF achieves better performance than LOF.}
	\label{fig:linematching}
\end{figure}

Table \ref{tab:PLVIORMSE} lists the root mean square error (RMSE) of the estimated trajectories by PL-VIO with different line tracking methods, where G-LOF represents LOF with prediction step assisted by gyroscope measurements. From the result, we can find that PL-VIO with LOF achieves better performance than that of LBD in most of the sequences. The line tracking length of LOF is longer than that of LBD, so PL-VIO system with LOF can better exploit the geometric information of environment to achieve better localization accuracy. Then, PL-VIO with G-LOF achieves better performance than PL-VIO with LOF in some sequences, which attributes to the prediction process in G-LOF. By using gyroscope measurements to predict a better initial guess in new image, the possibility that line alignment converges to the real one is increased. The comparison of line tracking results of LOF and G-LOF is shown in Fig. \ref{fig:linematching}.

\section{CONCLUSIONS}
In this paper, we propose a novel structure-aware Line tracking algorithm based entirely on Optical Flow (LOF). LOF employs the proposed gradient-based strategy to sample points on line that are appropriate for line optical flow calculation, and then tracks lines by performing the proposed two-step structure-aware line segment alignment method. In this way, the structural relationship between sampled points on line is fully utilized in the process of line segment alignment, and the influence of sampled points on occluding objectes is removed effectively. Then in order to eliminate the line tracking error accumulation and maintain the complete lines, LOF utilizes the intensity information to refine the orientation, position, and endpoints of the tracked lines. The proposed LOF algorithm is tested on two public datasets. The experiments show that LOF outperforms the state-of-the-art level in line tracking accuracy, length, and efficiency. Then, LOF can deal with partial occlusion and maintain the stable line endpoints. In addition, LOF improves the localization accuracy of VO system with lines by providing the better line tracking results. In future, we will study how to remove the line tracking outliers effectively.

\addtolength{\textheight}{-12cm}   



%
%
%


\bibliographystyle{IEEEtran}
\bibliography{LOF}
\end{document}